%% file: main.tex
\documentclass[letterpaper]{article} 
\usepackage{aaai2026}  
\usepackage{times}  
\usepackage{helvet}  
\usepackage{courier}  
\usepackage[hyphens]{url}  
\usepackage{graphicx} 
\usepackage{natbib}  
\usepackage{caption} 
\usepackage{algorithm}
\usepackage{algorithmic}

\usepackage{newfloat}
\usepackage{listings}
\DeclareCaptionStyle{ruled}{labelfont=normalfont,labelsep=colon,strut=off} 
\floatstyle{ruled}
\newfloat{listing}{tb}{lst}{}
\floatname{listing}{Listing}
\usepackage{multirow}
\usepackage{pifont}
\usepackage{xcolor} 
\usepackage{colortbl}
\usepackage{tcolorbox}
\usepackage{bbding}
\usepackage{amsmath}
\usepackage{graphicx}
\usepackage{amsmath}
\usepackage{amssymb}
\usepackage{float}
\newcommand{\dec}[1]{{\small \color[HTML]{EA4335} {(-#1)}}}
\newcommand{\imp}[1]{{\small \color[HTML]{34A853} {(+#1)}}}
\newtcolorbox{prompt}[1]{colback=gray!5,colframe=gray!35!black,fonttitle=\bfseries, title={#1}}
\definecolor{antiquebrass}{rgb}{0.8, 0.58, 0.46}

\definecolor{gray}{gray}{0.9}

\usepackage{makecell}
\usepackage{appendix}
\usepackage{algorithm}
\usepackage{algorithmic}
\usepackage{pifont}
\usepackage{enumitem}
\usepackage{newfloat}
\usepackage{listings}

\title{
\includegraphics[height=1cm]{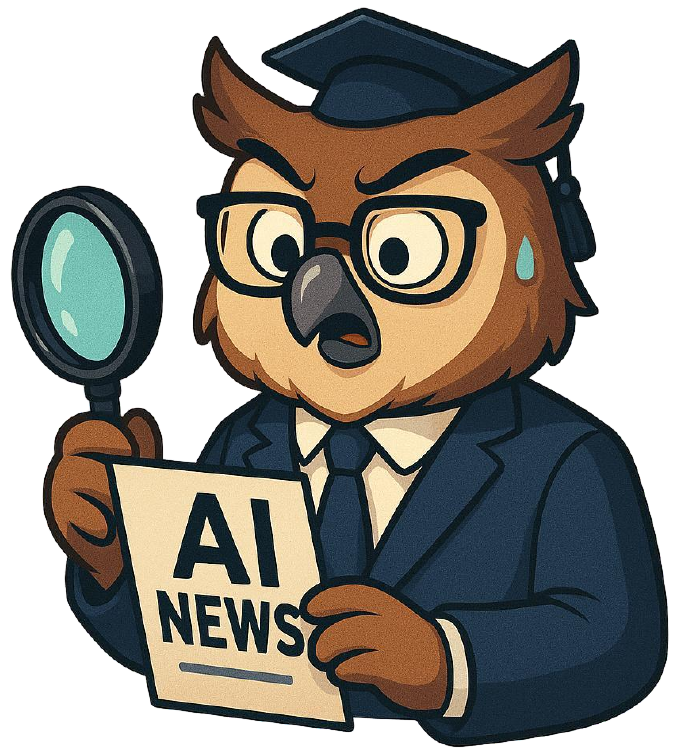} 
    \hspace{0.05cm} 
Drifting Away from Truth: GenAI-Driven News Diversity Challenges LVLM-Based Misinformation Detection}

\author{
    Fanxiao Li\textsuperscript{\rm 1},
    Jiaying Wu\textsuperscript{\rm 2}\thanks{Corresponding authors.},
    Tingchao Fu\textsuperscript{\rm 1}, Yunyun Dong\textsuperscript{\rm 3}, Bingbing Song\textsuperscript{\rm 3}, Wei Zhou\textsuperscript{\rm 4,5}$^{*}$
}
\affiliations{
         \textsuperscript{\rm 1} School of Information Science and Engineering, Yunnan University, \\ \textsuperscript{\rm 2} National University of Singapore,\\ \textsuperscript{\rm 3} School of Software and AI, Yunnan University, \\ \textsuperscript{\rm 4} Yunnan-Malaya Institute (School of Engineering), Yunnan University \\ \textsuperscript{\rm 5} State Key Laboratory of Vegetation Structure, Function and Construction (VegLab) \\
     lifanxiao@stu.ynu.edu.cn, jiayingw@nus.edu.sg, zwei@ynu.edu.cn
   
}

\usepackage{bibentry}

\begin{document}

\maketitle

\begin{abstract}
The proliferation of multimodal misinformation poses growing threats to public discourse and societal trust. While Large Vision-Language Models (LVLMs) have enabled recent progress in multimodal misinformation detection (MMD), the rise of generative AI (GenAI) tools introduces a new challenge: \textit{GenAI-driven news diversity}, characterized by highly varied and complex content. We show that this diversity induces \textit{multi-level drift}, comprising (1) \textit{model-level misperception drift}, where stylistic variations disrupt a model’s internal reasoning, and (2) \textit{evidence-level drift}, where expression diversity degrades the quality or relevance of retrieved external evidence. These drifts significantly degrade the robustness of current LVLM-based MMD systems. To systematically study this problem, we introduce \textsc{DriftBench}, a large-scale benchmark comprising 16,000 news instances across six categories of diversification. We design three evaluation tasks: (1) robustness of truth verification under multi-level drift; (2) susceptibility to adversarial evidence contamination generated by GenAI; and (3) analysis of reasoning consistency across diverse inputs. Experiments with six state-of-the-art LVLM-based detectors show substantial performance drops (average F1 $\downarrow$ 14.8\%) and increasingly unstable reasoning traces, with even more severe failures under adversarial evidence injection. Our findings uncover fundamental vulnerabilities in existing MMD systems and suggest an urgent need for more resilient approaches in the GenAI era. 

\end{abstract}


\section{Introduction}

Multimodal misinformation, often in the form of image-text combinations, poses escalating threats to public discourse, societal trust, and civic stability \cite{bovet2019influence, murayama2021dataset, wang2024harmfully, deceptiondecoder}. To counter this threat, Large Vision-Language Models (LVLMs) \cite{DBLP:journals/corr/gpt-4o, DBLP:journals/corr/Gemini, llava, meta2024llama} has emerged as a dominant approach for multimodal misinformation detection (MMD),  offering strong multimodal reasoning and retrieval-augmented verification capabilities \cite{kangur2025multireflect, qi2024sniffer, braun2024defame, li2025imrrf, geng2024multimodal, wu2025beyond}.


\begin{figure}[t]
\begin{center}
    \includegraphics[width=\linewidth]  {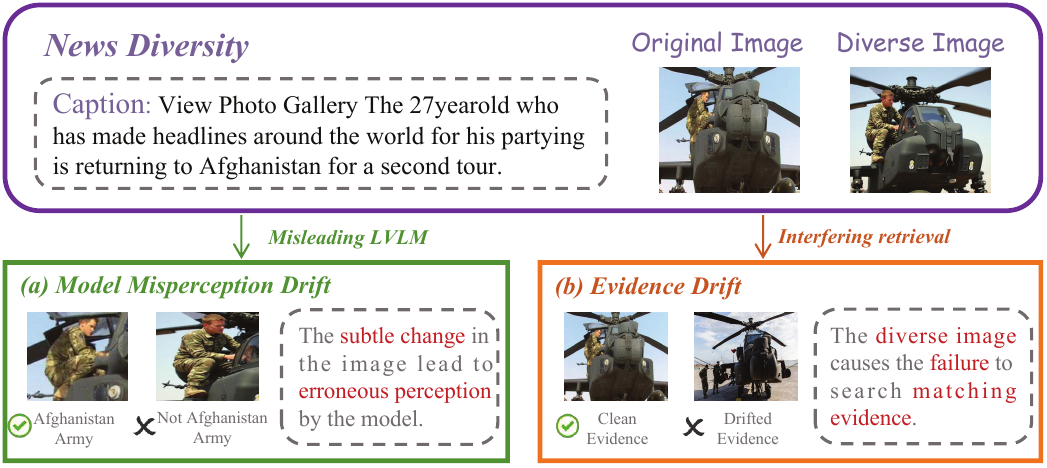}
    \caption{Illustration of \textbf{multi-level drift} induced by \textbf{GenAI-driven news diversity}, comprising (1) model-level misperception drift and (2) evidence-level drift.}
    \label{fig:drift_example}
\end{center}
\end{figure}

However, the rise of generative AI (GenAI) systems such as GPT-4o \cite{DBLP:journals/corr/gpt-4o} is reshaping the information landscape. GenAI enables the creation of highly diverse and stylistically varied content at scale \cite{kiskola2025generative, davenport2023generative, nishal2024envisioning, kieslich2024anticipating}, fundamentally transforming how news is composed, visualized, and disseminated. We refer to this phenomenon as \textbf{GenAI-driven News Diversity}, which manifests in two key forms: (1) \textit{Controlled News Diversity}, where users produce semantically consistent variants of the same news content through rewording, paraphrasing, or shifts in visual framing; and (2) \textit{Open-ended News Diversity}, where GenAI generates entirely novel content via text-to-image synthesis or narrative reconstruction.

While this diversity enhances expressiveness and audience reach, it also introduces substantial challenges for LVLM-based MMD systems. Most existing models either assess image-text coherence using internal knowledge \cite{wang2024mfc} or retrieve external evidence for fact verification \cite{li2025cmie, qi2024sniffer}. Both approaches become fragile under content variation, which gives rise to what we term \textbf{multi-level drift}. As illustrated in Figure~\ref{fig:drift_example}, we identify two key drift phenomena: (1) \textbf{Model-Level Misperception Drift}: Variations in surface form, including lexical, syntactic, or visual aspects, may distort the model’s perception, even when semantics remain unchanged. This issue is exacerbated by LVLMs’ known hallucination tendencies \cite{bai2024hallucination, liu2024hallucination}. (2) \textbf{Evidence-Level Drift}: Content diversity compromises evidence retrieval by reducing the likelihood of finding exact matches, leading to loosely relevant or misleading external evidence \cite{chung2023guard, xiao2021you, feng2020adversarial}.
Beyond natural drift, GenAI also enables adversarial manipulation through \textbf{guided retrieval attacks}, wherein malicious actors flood the web with strategically crafted synthetic content. Such contamination can bias retrieval pipelines, poisoning the fact-checking process and degrading LVLM performance \cite{xu2024invisible, dai2023llms}. Despite their growing risk, these vulnerabilities remain insufficiently studied.

To bridge this gap, we conduct a systematic investigation into how GenAI-driven news diversity affects LVLM-based misinformation detection. We pose three core research questions: (1) How does GenAI-driven news diversity affect the robustness of current LVLM-based MMD systems? (2) To what extent does adversarial evidence contamination degrade factual verification via guided retrieval of fabricated evidence? (3) How does content diversity influence LVLMs’ internal reasoning behavior during fact-checking?

To this end, we introduce \textsc{DriftBench}, a large-scale benchmark consisting of 16,000 news instances spanning six diversified categories across both image and text modalities, with human validation to ensure semantic fidelity. Specifically, we source high-quality real and fake samples from NewsCLIPpings \cite{newsclippings} and systematically diversify them using an automated GenAI-based pipeline grounded in our news diversity taxonomy. For real news, we apply \textit{controlled} diversification to the image, text, or both while preserving factual consistency. For fake news, we incorporate both \textit{controlled} and \textit{open-ended} transformations, rewriting text to fabricate false narratives and generating misleading images aligned with the fabricated content. The resulting dataset includes eight structured categories, consisting of six diversified variants and two original classes, thereby enabling rigorous evaluation of LVLM robustness under GenAI-driven content variation.

With \textsc{DriftBench}, we design three corresponding evaluation tasks: (1) \textbf{Performance under Diversity}: Evaluate how SOTA LVLM-based detectors handle controlled and open-ended content variation, focusing on degradation due to model- and evidence-level drift.
(2) \textbf{Robustness to Malicious Evidence Contamination}: Simulate guided retrieval attacks by injecting GenAI-generated misleading evidence into the verification pipeline. (3) \textbf{Reasoning Behavior Analysis}: Employ an LVLM-as-a-Judge framework \cite{li2024judge-1, chen2024judge-2, pu2025judge-3} to assess models’ explanation quality and attribution patterns under diverse input conditions.

Our empirical analysis covers both general-purpose LVLMs and task-specific detectors. Results show that GenAI-driven diversity leads to significant performance drops across all models (average F1 $\downarrow$ 14.8\%), with malicious evidence causing further degradation. Notably, we observe that real and fake content are differently affected by diversity; vanilla and task-specific models exhibit divergent sensitivities; and diverse inputs distort reasoning trajectories in non-trivial ways. Our findings reveal fundamental limitations in current LVLM-based detection pipelines and lay the groundwork for designing robust, diversity-aware MMD systems to accommodate GenAI-driven news diversity.

\section{Related Work}

\input{table/benchmark_compare}

\subsection{LVLM-Based MMD Approaches}
LVLM-based multimodal fact-checking is the current mainstream method for multimodal misinformation detection.
SNIFFER \cite{qi2024sniffer} performs two-stage fine-tuning based on InstructBLIP \cite{dai2023instructblip}, aiming to detect misinformation from the perspective of entity matching and evidence verification. CMIE \cite{li2025cmie} and E2LVLM \cite{wu2025e2lvlm} further explore how to discover deeper relationships and optimize the selection and usage of external evidence. LRQ-FACT \cite{lrq-fact} and LEMMA \cite{xuan2024lemma} optimize retrieval strategies to acquire more targeted external evidence.
These methods fully investigate the capabilities of LVLMs in retrieval, comprehension, and reasoning, and are optimized around the acquisition and utilization of high-quality evidence. However, most of them assume that the input claim is static and expressed in a single form, lacking systematic investigation and evaluation of such methods' performance under news diversity and the resulting multi-level drift.

\subsection{Multimodal Misinformation Benchmarks}
To evaluate multimodal misinformation detection (MMD), several benchmarks have been introduced. DGM4 \cite{shao2023dgm4} targets fine-grained visual and textual manipulations. NewsCLIPpings \cite{newsclippings} constructs out-of-context (OOC) cases by mismatching images and text with external evidence. VERITE \cite{papadopoulos2024verite} collects real-world OOC misinformation from fact-checking platforms to analyze modality bias. MiRAGenNews \cite{MiRAGeNews} employs GenAI techniques to synthesize misinformation image–text pairs, while MMFakeBench \cite{liu2024mmfakebench} aggregates misinformation from both real and GenAI sources to assess robustness under mixed conditions. MFC-Bench \cite{wang2024mfc} probes LVLMs’ internal fact-checking capabilities.

In contrast, \textsc{DriftBench} is the first benchmark to systematically examine challenges introduced by \textbf{GenAI-driven news diversity}, with a focus on two overlooked forms of drift: \textbf{model misperception drift} and \textbf{evidence drift}. By generating semantically diverse real and fabricated variants through both controlled and open-ended transformations, it addresses a critical gap in existing benchmarks. A comparative overview is provided in Table~\ref{benchmark_compare}.

\section{\textsc{DriftBench}}

\begin{figure*}[t]
\begin{center}
    \includegraphics[width=\linewidth]  {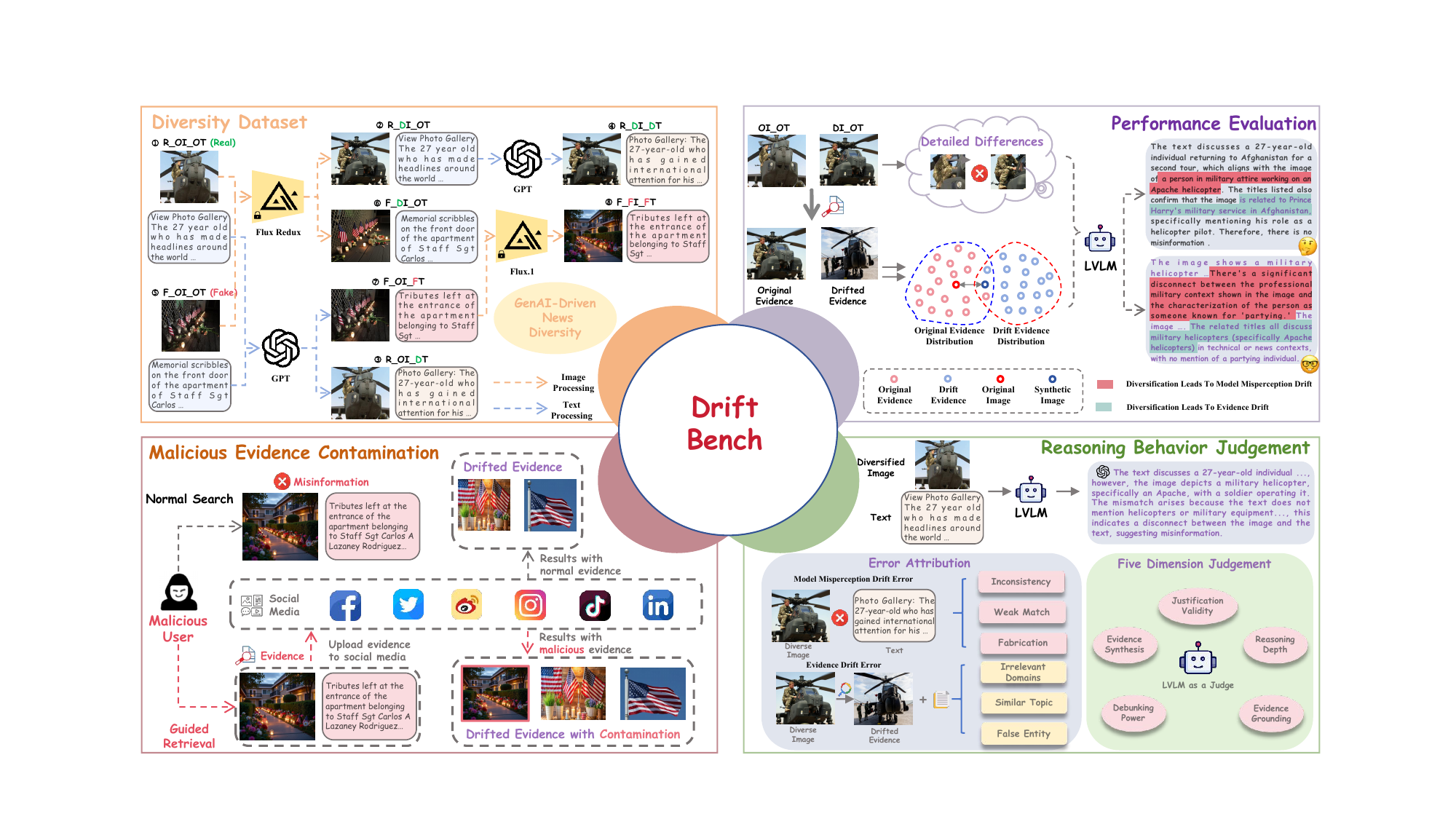}
    \caption{Overview of \textsc{DriftBench}, illustrating the construction of diversified news variants and the three evaluation tasks.}
    \label{fig:framework}
\end{center}
\end{figure*}

To systematically evaluate how \textit{GenAI-driven news diversity} induces multi-level drift in LVLM-based multimodal misinformation detection, we introduce \textsc{DriftBench}, a large-scale benchmark of 16,000 news instances across eight categories. These categories comprise two base types (real and fake news with human-written text and authentic images) and six diversification strategies. Built through an automated GenAI-based synthesis pipeline applied to real-world news, \textsc{DriftBench} enables scalable, controlled, and semantically aligned diversification. Each instance contains an image–text pair—either original or diversified—together with externally retrieved web evidence. An overview of the dataset construction and evaluation setup is shown in Figure~\ref{fig:framework}. Our code, data, and supplementary materials are publicly available\footnote{https://github.com/fanxiao15/DriftBench}.

\subsection{Taxonomy of GenAI-Driven News Diversity}
\label{taxonomy}

To comprehensively assess the impact of content variation on multimodal misinformation detection, we propose a taxonomy encompassing eight types of image-text instances. These are defined along two orthogonal dimensions: \textbf{(1) news authenticity} (\textit{Real} vs. \textit{Fake}) and \textbf{(2) diversification strategy} (\textit{Controlled} vs. \textit{Open-Ended}). Controlled diversity involves deliberate and precise modifications to image or text while preserving semantic consistency. Open-ended diversity allows flexible, generative transformations, often introducing false or misleading information, thereby increasing uncertainty.

We begin with two foundational types that serve as the seed for diversification:

\begin{itemize}[leftmargin=*]
    \item \textbf{R\_OI\_OT}: A real news item sourced from a reputable outlet, composed of an authentic image (\textit{OI}) and a human-written text (\textit{OT}) that are semantically aligned.
    \item \textbf{F\_OI\_OT}: An out-of-context (OOC) misinformation instance \cite{newsclippings, qi2024sniffer}, created by mismatching the image and text from two different but individually trustworthy news reports—misleading when paired together.
\end{itemize}

Building on these base types, we define six variant categories by modifying the image, text, or both using controlled or open-ended GenAI techniques:

\begin{itemize}[leftmargin=*]
    \item \textbf{R\_DI\_OT}: A real news instance where the image is diversified using GenAI (\textit{DI}), while the original text remains unchanged.
    \item \textbf{R\_OI\_DT}: A real news instance where the text is diversified (\textit{DT}) while preserving the original image.
    \item \textbf{R\_DI\_DT}: A real instance where both the image and text are diversified in a semantically faithful manner.
    \item \textbf{F\_DI\_OT}: An OOC fake instance where the image is replaced by a GenAI-diversified version, while keeping the mismatched original text.
    \item \textbf{F\_OI\_FT}: A fake instance where the original image is paired with a GenAI-fabricated text (\textit{FT}) that conveys a false narrative.
    \item \textbf{F\_FI\_FT}: A fully fabricated fake instance where both the image (\textit{FI}) and text (\textit{FT}) are GenAI-synthesized.
\end{itemize}

This taxonomy enables a systematic evaluation of how controlled and open-ended diversification strategies, applied to both real and fake instances, influence the robustness and reasoning behavior of LVLM-based misinformation detection systems.

\subsection{Dataset Creation}
\label{sec: data_creation}

Based on the proposed taxonomy, we construct \textsc{DriftBench}, a large-scale benchmark comprising real and fake news instances systematically diversified through GenAI-driven content variation.
We begin by sourcing 4,000 high-quality samples (2,000 real and 2,000 fake) from the NewsCLIPpings dataset \cite{newsclippings}. 
Real instances comprise semantically aligned image-caption pairs published by reputable news outlets. Fake instances are constructed via OOC manipulation, where the image and text are drawn from two different trustworthy reports. While individually accurate, their combination results in a misleading narrative with high surface plausibility.

We apply different transformation strategies based on the type of instance. For real samples, we apply only \textit{controlled} diversification to preserve factual integrity. For fake samples, we introduce both controlled and \textit{open-ended} diversity to simulate more flexible or fabricated misinformation.

Text diversifications and fabrications are generated using GPT-4o \cite{DBLP:journals/corr/gpt-4o}. For text diversification (\textit{DT}), we prompt the model to rephrase the original text while preserving its semantics. For text fabrication (\textit{FT}), we prompt the model to generate a false narrative by altering the described event.
Image diversifications (\textit{DI}) are generated using \textit{FLUX.1 Redux [dev]} (an open-source image variant generation model) \cite{labs2025flux1kontextflowmatching}, the model generates visually varied but semantically consistent depictions of the original content. Image fabrications (\textit{FI}) are generated using \textit{FLUX.1 [dev]} (an open-source text-to-image model) \cite{flux2024}, we condition the image generation on the fabricated narrative \textit{FT}, yielding fully synthetic misinformation instances.

Implementation details and prompt templates are provided in Appendix~\ref{app: Dataset_Creation_Details}.

\paragraph{Evidence Retrieval.}
For each image-text pair, we retrieve external evidence to support LVLM-based multimodal fact-checking.
For image-based retrieval, we use a web crawler to identify webpages containing the image and extract their page titles as evidence.
For text-based retrieval, we adopt the query generation strategy from LEMMA \cite{xuan2024lemma}, using the Serper API to retrieve relevant web documents.

\input{table/main_result}

\subsection{Data Quality Assessment}
\label{sec: human_eval}

Preserving semantic meaning is essential for high-quality news diversification. To assess the quality of the diversified instances in \textsc{DriftBench}, we conducted a human evaluation to determine whether the core meaning of the original content is retained after diversification. Since the \textit{OI\_FT} and \textit{FI\_FT} types involve fake news with fabricated content, they represent open-ended generation and lack a verifiable semantic reference. Therefore, we restrict our evaluation to the four types of diversified \textit{real} instances, where semantic fidelity can be reliably assessed.

We randomly sampled 40 image-text pairs from each diversification type, yielding a total of 160 diversified instances. Three graduate students served as annotators. Each annotator independently reviewed all 160 original-diversified pairs and provided a binary judgment on whether the diversified version preserved the core meaning of the original. Annotation guidelines and examples are detailed in Appendix~\ref{app: human_eval}.

We report two metrics to assess data quality: (1) semantic preservation accuracy, computed from aggregated human annotations, and (2) inter-annotator agreement using Fleiss' $\kappa$ \cite{fleiss1971measuring}. Results show that \textsc{DriftBench} exhibits high semantic consistency and labeling reliability, with 98.6\% accuracy and a Fleiss' $\kappa$ of 0.872 for real-instance variants, and 96.6\% accuracy with a Fleiss' $\kappa$ of 0.741 for controlled fake variants.




\subsection{Evaluation Tasks and Metrics}
\label{sec: eval_task_setup}

We design three evaluation tasks to assess:
(1) performance degradation under \textit{model-level misperception drift} and \textit{evidence-level drift} introduced by news diversity;
(2) robustness to \textit{malicious evidence contamination}; and
(3) the influence of news diversity on \textit{reasoning behavior and explanation quality}.

\paragraph{Task 1: Performance Analysis under News Diversity.}
We evaluate how both Vanilla LVLMs and task-specific MMD models perform on diversified instances in \textsc{DriftBench}, measuring the impact of semantic-preserving (controlled) and misleading (open-ended) variations on the models' MMD performance.

\paragraph{Task 2: Robustness to Malicious Evidence Contamination.} To simulate guided retrieval attacks, we inject adversarially generated evidence (i.e., crafted to falsely support or refute the claim) into the evidence set. This mimics real-world scenarios where search engines are manipulated by fabricated content. We evaluate models' robustness by measuring performance drop before and after evidence poisoning, isolating the sensitivity of each model to retrieval contamination. The generation procedure and insertion strategy are detailed in Appendix~\ref{app: malicious}.

\paragraph{Task 3: Reasoning Behavior Analysis.}
We adopt an LVLM-as-Judge evaluation framework to analyze model reasoning under drift. This includes:
\begin{itemize}[leftmargin=*]
\item \textbf{Error Attribution:} We manually label failure cases according to two error types: (1) model-level misperception drift and (2) evidence-level drift. A detailed taxonomy is provided in Appendix~\ref{app: error_attribution}.
\item \textbf{Explanation Evaluation:} We assess the quality of model-generated explanations across five dimensions: justification validity, evidence grounding, evidence synthesis, reasoning depth, and debunking power. Full guidelines and scoring rubrics are included in Appendix~\ref{app: explanation_eval}.
\end{itemize}

Following established practices in LVLM-based MMD evaluation \cite{li2025cmie,xuan2024lemma,qi2024sniffer}, we report Accuracy and F1 as the primary metrics for detection performance.





\section{Experiments}

\subsection{Experimental Setup}

Using \textsc{DriftBench}, we conduct a comprehensive evaluation of both representative Vanilla LVLMs and state-of-the-art LVLM-based MMD approaches.

\noindent\textbf{Vanilla LVLMs.} We assess three widely used models: (1) \textbf{GPT-4o-mini} \cite{DBLP:journals/corr/gpt-4o}: a proprietary large vision-language model. (2) \textbf{Claude-3.7-Sonnet} \cite{claude3_7}: a large reasoning model with multimodal capabilities. (3) \textbf{Qwen-VL-Plus} \cite{bai2023qwen-vl}: a strong open-source vision-language model. Prompting strategies and instruction templates used for Vanilla LVLMs are detailed in Appendix (Figure \ref{appfig:LVLMs-prompt}).

\noindent\textbf{LVLM-Based MMD Methods.} We also evaluate three state-of-the-art MMD approaches. (1) \textbf{SNIFFER} \cite{qi2024sniffer}, which fine-tunes LVLMs and retrieves external evidence directly through image-based querying. (2) \textbf{CMIE} \cite{li2025cmie}, which models deep semantic alignment between image and text while mitigating noise in retrieved evidence. and (3) \textbf{LEMMA} \cite{xuan2024lemma}, which employs a modular pipeline that first generates optimized queries and then retrieves high-quality evidence.


\subsection{Performance under GenAI-Driven Diversity}
\label{sec: Performance_under_GenAI-Driven_Diversity}
\textbf{\textit{Multimodal LVLMs exhibit limited generalization under news diversity, with asymmetric performance between real and fake Instance.}}
As shown in Table~\ref{tab: main_results}, GenAI-driven news diversity causes significant performance degradation across all evaluated models. This demonstrates that even controlled, semantically consistent perturbations can severely compromise the reliability of current LVLM-based detectors. Multi-level drift induces a bias toward overpredicting ``fake''. For real instances, both precision and recall drop substantially due to a combination of model-level misperception drift (e.g., hallucinated misalignment between modalities) and evidence-level drift (e.g., irrelevant or misleading retrieval). For fake instances, while recall improves, precision declines. This suggests that the models become overconfident in detecting misinformation, even when uncertain.

\textbf{\textit{Image diversity has a stronger impact than text diversity, and open-ended diversity is more disruptive than controlled diversity.}} 
\begin{figure}[h]
\begin{center}
    \includegraphics[width=1\linewidth]  {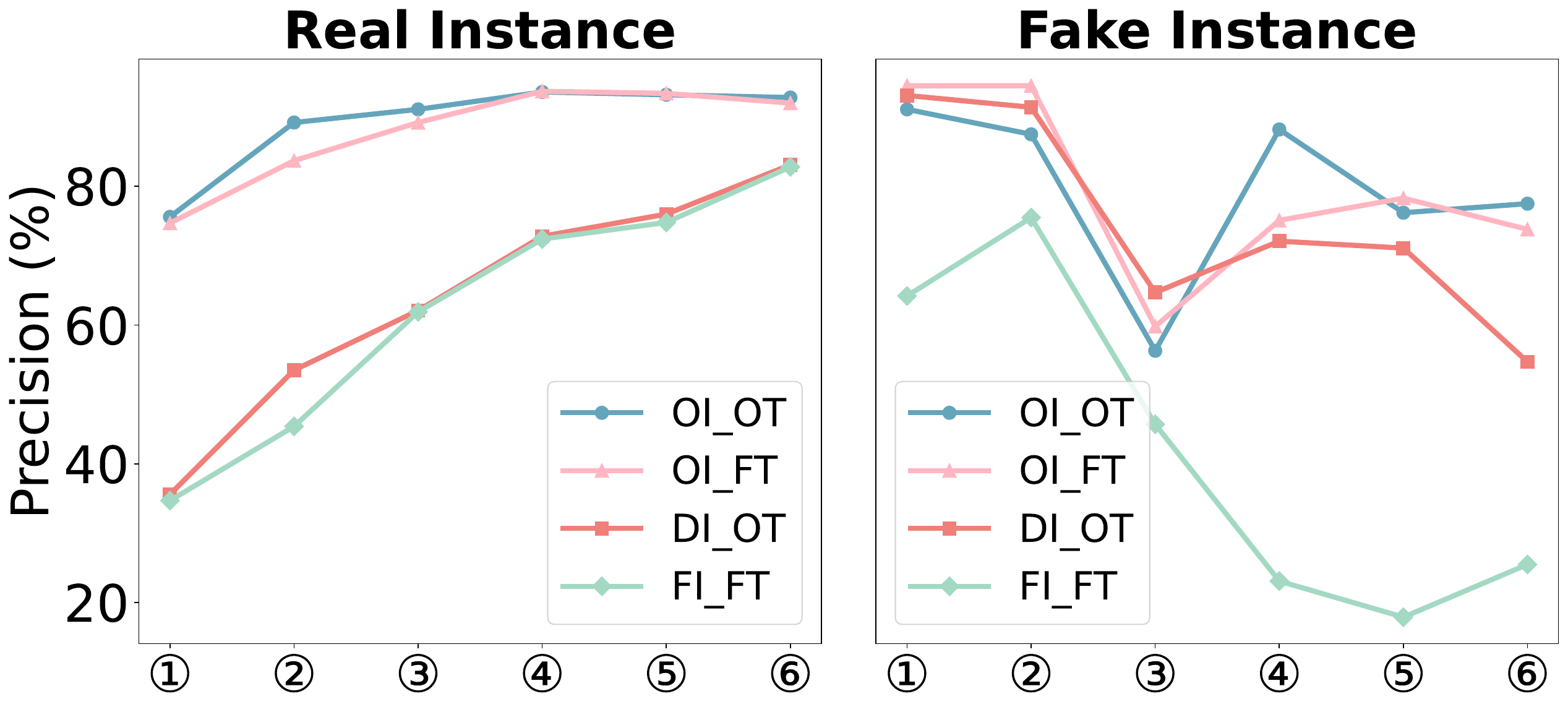}
    \caption{Precision (\%) of representative LVLM-based MMD approaches (\ding{172}: \textit{GPT-4o-mini}; \ding{173}: \textit{Claude-3.7}; \ding{174}: \textit{Qwen-VL}; \ding{175}: \textit{CMIE}; \ding{176}: \textit{Sniffer}; \ding{177}: \textit{Lemma}) under all types of \textit{controlled} and \textit{open-ended} news diversity.}
    \label{fig:diff_type_recall}
\end{center}
\end{figure}
As shown in Figure~\ref{fig:diff_type_recall}, text diversification (\textit{OI\_DT}) has minimal impact, while performance declines are primarily driven by image diversification (\textit{DI\_OT}), detailed results in Appendix \ref{app: detailed_results}.
Notably, open-ended diversity leads to greater performance drops than controlled diversity. In these cases, the generated images may be perceived as too well-aligned with the fabricated text, misleading the model into incorrectly classifying the instance as real. This highlights a fundamental challenge posed by GenAI-generated misinformation, where visual-textual coherence can be artificially inflated.

\begin{figure}[h]
\begin{center}
    \includegraphics[width=1\linewidth]  {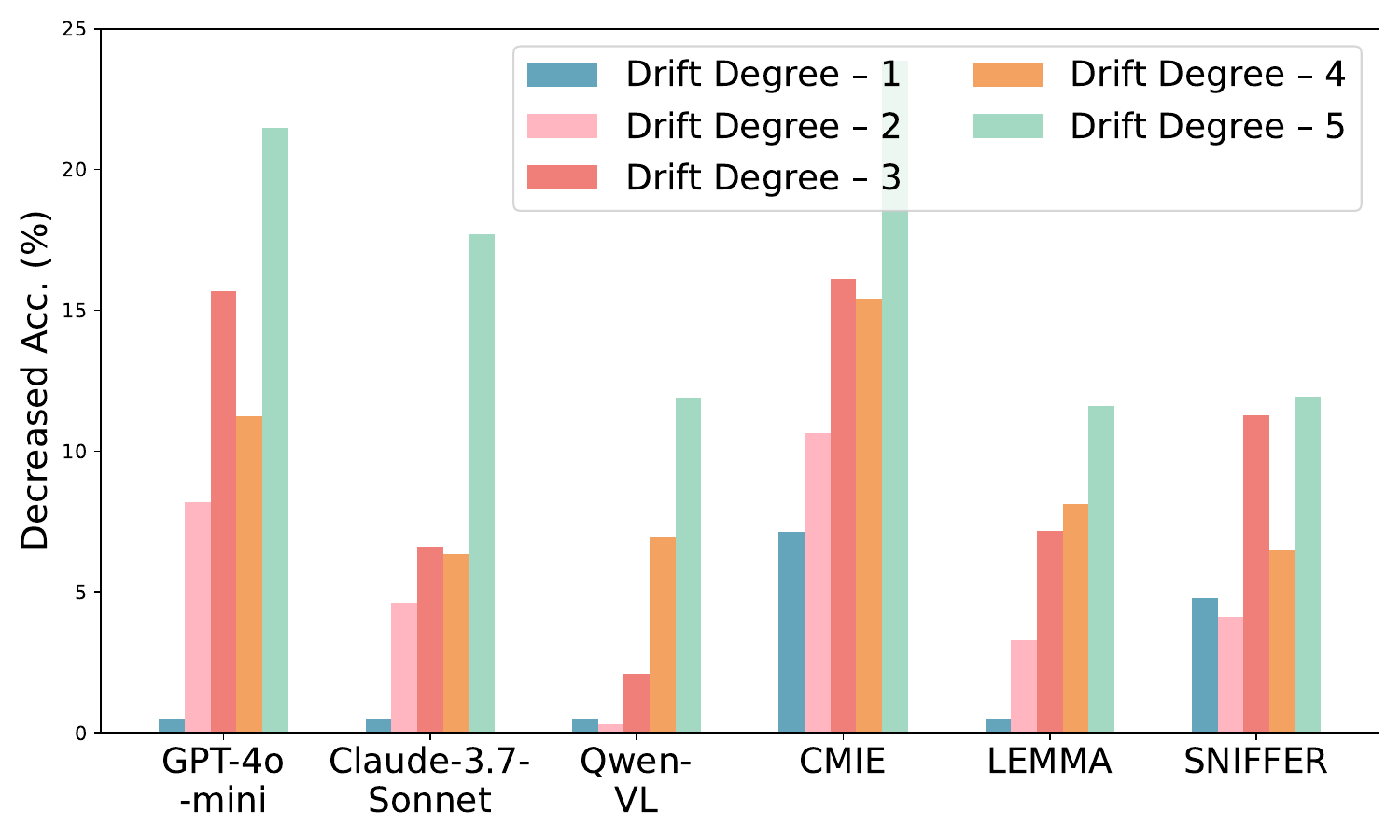}
    \caption{Accuracy degradation of MMD methods under increasing evidence drift severity in the \textit{DI\_OT} setting. Higher drift degrees lead to more pronounced performance drops.}
    \label{fig:acc_drift_degree}
\end{center}
\end{figure}

\textbf{\textit{Model performance degrades more sharply with increasing evidence drift severity.}}
Figure~\ref{fig:acc_drift_degree} illustrates how detection accuracy declines as the degree of evidence drift increases, ranging from mild (Degree 1) to severe (Degree 5). While the degradation is not strictly linear, the trend is clear: performance steadily worsens from low to moderate to high drift levels. As drift severity grows, the retrieved evidence becomes increasingly misaligned or ambiguous, making it harder for models to ground their predictions accurately. These findings highlight the urgent need for drift-aware verification mechanisms that can remain robust in the face of evidence drifts induced by GenAI-driven diversity.

\subsection{Robustness to Evidence Contamination}
\label{sec: robustness_to_evidence_contamination}

\textbf{\textit{Contaminated evidence further degrades model performance and can steer predictions in misleading directions.}}  As discussed in Section~\ref{sec: eval_task_setup}, malicious actors can manipulate retrieval pipelines by injecting fabricated or misleading content into the evidence set. Table~\ref{tab: pollution} shows that introducing such adversarial evidence causes a clear performance drop across all evaluated methods. Detailed results in Appendix~\ref{app: detailed_results} indicate that real instances are increasingly misclassified as fake when contaminated evidence is present. For fake instances, the misleading evidence often supports the false narrative, leading models to mistakenly classify them as real and thus producing the outcome intended by the adversary.

Notably, Vanilla LVLMs show greater performance fluctuation under contamination, suggesting they are more vulnerable to evidence interference in noisy environments. In comparison, task-specific MMD approaches such as CMIE and LEMMA maintain more stable F1 scores, due to their dedicated mechanisms for filtering and interpreting evidence. However, their performance also declines when exposed to adversarial content. This highlights the critical need for retrieval-aware defenses that can detect and mitigate the effects of evidence manipulation in LVLM-based misinformation detection systems.

\input{table/pollution_simple}

\subsection{Model Reasoning Behavior Analysis}
\begin{figure*}[h]
\begin{center}
    \includegraphics[width=1\linewidth]  {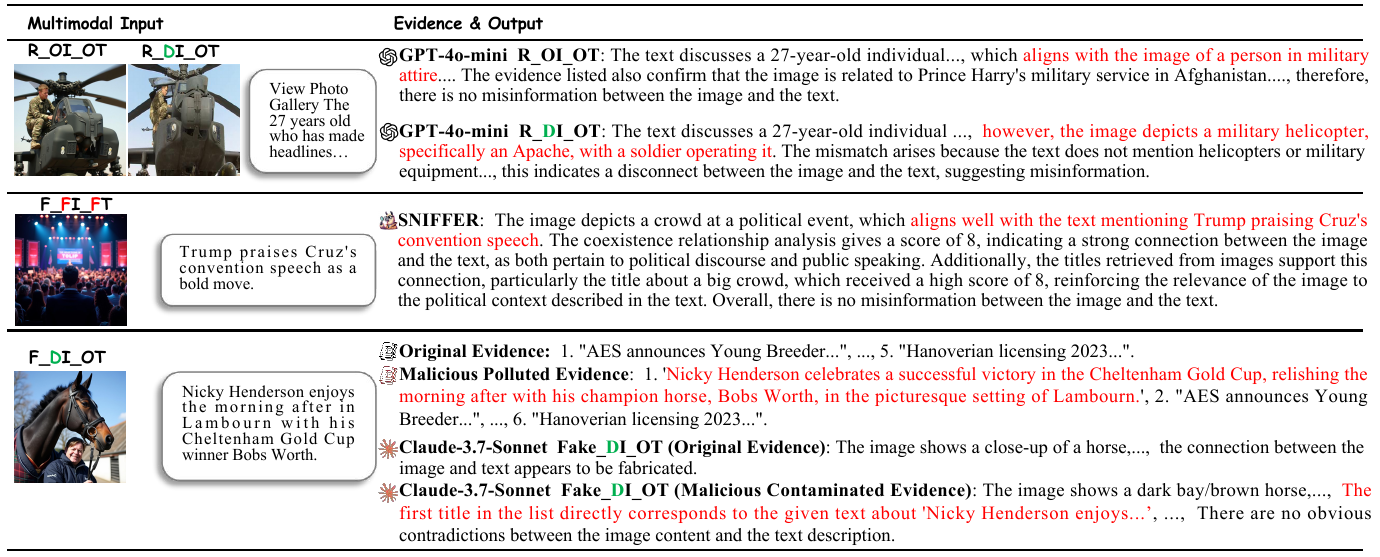}
    \caption{Reasoning explanations generated by LVLMs under controlled settings (R\_OI\_OT, R\_DI\_OT), open ended scenarios (F\_FI\_FT), and cases involving malicious evidence contamination (F\_DI\_OT).}
    \label{fig:case_study}
\end{center}
\end{figure*}

\begin{figure}[h]
\begin{center}
    \includegraphics[width=1\linewidth]  {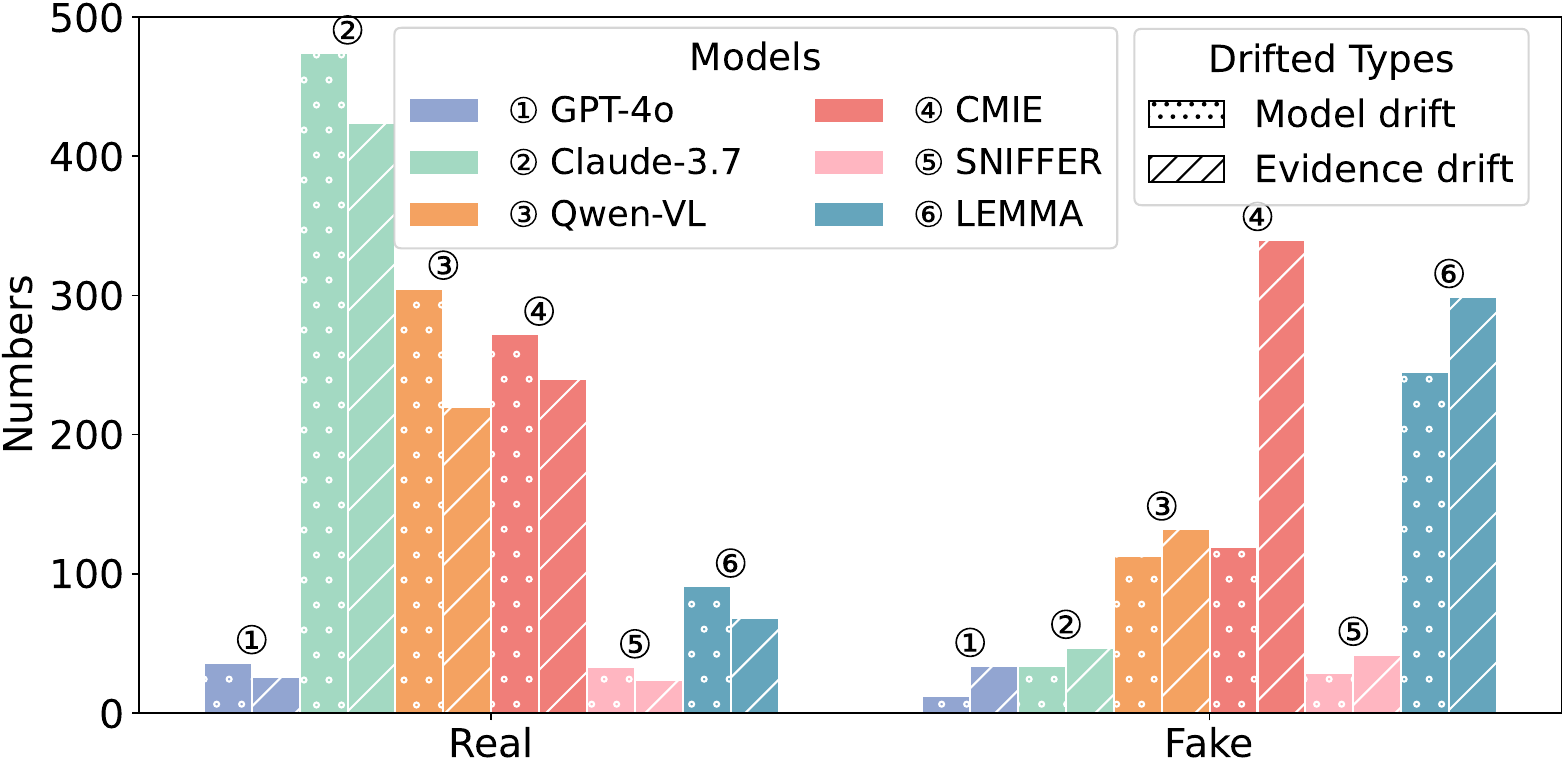}
    \caption{Error attribution showing the distribution of error types across different methods for real and fake instances.}
    \label{fig:error_attribution}
\end{center}
\end{figure}

\textbf{\textit{Visual-textual consistency influences model misclassification tendencies across real and fake instances.}} We analyze error attribution across diversified instances, as illustrated in Figure~\ref{fig:error_attribution}. While final misjudgments often result from both model misperception and evidence drift, their relative contributions differ between real and fake news.
For real instances, errors are more frequently caused by model-level misperception drift, suggesting that when image-text consistency remains strong, models tend to rely on this internal alignment for prediction. In contrast, for fake instances, errors are more often attributed to evidence-level drift, indicating that once image-text alignment is questioned, LVLMs tend to shift their focus to external evidence for verification.

\textbf{\textit{Vanilla LVLMs and MMD-specific models show distinct reasoning behaviors under news diversity.}}
As shown in Figure~\ref{fig:diff_type_recall}, Vanilla LVLMs are more sensitive to variation in real instances, while MMD-specific models exhibit greater robustness due to their dedicated vision-text alignment modules. However, for fake instances, MMD-specific models experience larger performance drops, primarily due to their reliance on external evidence, which is more vulnerable to drift. This highlights the dual challenge: LVLMs are perceptually fragile, while retrieval-based models are susceptible to evidence contamination.


\textbf{\textit{Different news diversity types affect explanation quality and reasoning dynamics in distinct ways.}}
We report explanation evaluation results across five dimensions in Appendix~\ref{app: explanation_eval}. Text-based diversification typically yields better explanation quality, potentially because LLMs favor content that resembles their generative style. Open-source LVLMs show more instability, while reasoning-focused models maintain consistent explanation quality but still fall short in accuracy due to lacking specialized modules.

Under diversification, Vanilla LVLMs exhibit weakened evidence integration yet maintain relatively stable reasoning depth, often misusing evidence without full synthesis. In contrast, task-specific MMD baselines show a correlated decline in both evidence usage and reasoning depth, revealing their dependency on structured reasoning components. These findings underscore the need for more adaptive and drift-resilient reasoning mechanisms in the face of GenAI-driven variation.

\subsection{Case Study}
In Figure \ref{fig:case_study}, we demonstrate the impact of different types of news diversity and malicious evidence contamination on LVLMs' ability to detect misinformation. Controlled news diversity leads to the model's misperception drift, while open-ended diversity causes the model to perceive the image-text pair as well-aligned, resulting in misclassification. As for evidence contamination, the model's judgment is influenced by malicious evidence, steering predictions in misleading directions.

\section{Conclusion}
We present \textsc{DriftBench}, the first benchmark explicitly designed to evaluate the vulnerabilities of LVLM-based misinformation detection in the face of GenAI-driven news diversity. Our diversified benchmark demonstrates that both controlled and open-ended content variation, along with adversarial evidence contamination, can substantially weaken the performance of current LVLM-based MMD systems. Beyond accuracy degradation, our analysis reveals distinct reasoning behaviors and error patterns among different model classes, reflecting varied reliance on visual-textual coherence and external evidence. These findings emphasize the need for robust, diversity-aware strategies to ensure reliable misinformation detection. \textsc{DriftBench} provides a foundation for advancing trustworthy verification methods in an increasingly complex and manipulated information landscape.

\clearpage
\section{Acknowledgment}
This work is supported by the Yunnan Province expert workstations (Grant No. 202305AF150078), National Natural Science Foundation of China (Grant Nos. 62162067, 62562061, 62502422 and  62462067), Yunnan Fundamental Research Project (Grant Nos. 202401AT070474, 202501AU070059), Yunnan Province Special Project (Grant No.202403AP140021), Yunnan Provincial Department of Education Science Research Project (Grant Nos. 2025J0006, 2024J0010 and 2025J0007), Scientific Research and Innovation Project of Postgraduate Students in the Academic Degree of Yunnan University (KC-4248590) and China Scholarship Council (CSC) program. This research is also supported by the Ministry of Education, Singapore, under its MOE AcRF TIER 3 Grant (MOE-MOET32022-0001).

\bibliography{aaai2026}

\clearpage

\input{appendix}
\end{document}

%% file: table/benchmark_compare.tex
\begin{table*}[t]
\renewcommand\arraystretch{0.9}
\setlength{\tabcolsep}{0.9 pt} 
\small
  \begin{center}
    \caption{Comparison between \textsc{DRIFTBENCH} and prior benchmarks on multimodal misinformation detection.}
\begin{tabular}{cccccccc}

\hline 
\multirow{2}{*}{\textbf{Benchmark}} & \multirow{2}{*}{\textbf{Information Type}}  &  \multicolumn{2}{c}{\textbf{Modality}} & \multirow{2}{*}{\textbf{\makecell{Generated \\ Content}}}& \multirow{2}{*}{\textbf{\makecell{External\\ Evidence}}}& \multirow{2}{*}{\textbf{\makecell{News \\ Diversity}}}& \multirow{2}{*}{\textbf{Drift}} \\ \cline{3-4}
  & & \textbf{Textual} & \textbf{Visual} & &  & & \\ \hline
  DGM4 \cite{shao2023dgm4} &  \makecell{Multimedia manipulation}  & \ding{52} & \ding{52} & - & - & - & - \\
  NewsCLIPpings \cite{newsclippings} & OOC misinformation & \ding{52} & \ding{52} & - & \ding{52} & - & - \\
  VERITE \cite{papadopoulos2024verite} & Real-world misinformation & \ding{52} & \ding{52} & - & \ding{52} & - & - \\
  MMFakeBench  \cite{liu2024mmfakebench} & Multi-source misinformation & \ding{52} & \ding{52} & \ding{52} & - & - & - \\
  MFC-Bench \cite{wang2024mfc} & Multimodal fact-checking & \ding{52} & \ding{52} & - & - & - & - \\
  MiRAGenNews \cite{MiRAGeNews} & AI-generated misinformation & \ding{52} & \ding{52} & \ding{52} & - & - & - \\ \hline
  \textsc{DriftBench} (ours) & News diversity \& Drift & \ding{52} & \ding{52} & \ding{52} & \ding{52} & \ding{52} & \ding{52} \\ \hline

\end{tabular}
    \label{benchmark_compare}
  \end{center}
\end{table*}

%% file: table/main_result.tex
\begin{table*}[t]
\renewcommand\arraystretch{1}
\setlength{\tabcolsep}{3 pt} 
\small
  \begin{center}
    \caption{Performance comparison of LVLM-based multimodal misinformation detection methods under the \textit{DI\_OT} setting within Controlled News Diversity. Since \textit{DI\_OT} applies to both real and fake instances, we report results (in percentage) for this category as a representative example.}
\begin{tabular}{cclllllll}
\hline \hline
\multirow{2}{*}{\textbf{Infer Type}} & \multirow{2}{*}{\textbf{Data Type}} & \multirow{2}{*}{\textbf{Accuracy}} &  \multicolumn{3}{c}{\textbf{Real}} & \multicolumn{3}{c}{\textbf{Fake}} \\ \cline{4-9}
  & & & \multicolumn{1}{c}{\textbf{Precision}} & \multicolumn{1}{c}{\textbf{Recall}} & \multicolumn{1}{c}{\textbf{F1}} & \multicolumn{1}{c}{\textbf{Precision}} & \multicolumn{1}{c}{\textbf{Recall}} & \multicolumn{1}{c}{\textbf{F1}} \\ \hline

 \multirow{2}{*}{\textbf{GPT-4o-mini}} 
   & Realistic & 83.3 & 89.4 & 75.6 & 81.9 & 78.9 & 91.1 & 84.5 \\
   & Diversified & 64.4 \dec{18.9} & 83.7 \dec{6.0} & 35.7 \dec{39.9} & 50.0 \dec{31.9} & 59.1 \dec{19.8} & 93.1 \imp{2.0} & 72.3 \dec{12.2} \\ \hline

 \multirow{2}{*}{\textbf{Claude-3.7-Sonnet}} 
   & Realistic & 88.3 & 87.7 & 89.2 & 88.4 & 89.0 & 87.5 & 88.2 \\
   & Diversified & 72.4 \dec{15.9} & 86.1 \dec{1.6} & 53.5 \dec{35.7} & 66.0 \dec{22.4} & 66.3 \dec{22.7} & 91.4 \imp{3.9} & 76.8 \dec{11.4} \\ \hline

 \multirow{2}{*}{\textbf{Qwen-VL}} 
   & Realistic & 73.7 & 67.6 & 91.1 & 77.6 & 86.3 & 56.3 & 68.2 \\
   & Diversified & 63.7 \dec{10.0} & 63.9 \dec{3.7} & 62.7 \dec{28.4} & 63.3 \dec{14.3} & 63.5 \dec{22.8} & 64.7 \imp{8.4} & 64.1 \dec{4.1} \\ \hline

 \multirow{2}{*}{\textbf{CMIE}} 
   & Realistic & 90.9 & 88.8 & 93.6 & 91.1 & 93.2 & 88.2 & 90.6 \\
   & Diversified & 72.4 \dec{18.5} & 72.2\dec{16.6} & 72.8\dec{20.8} & 72.5\dec{18.6} & 72.6\dec{20.6} & 72.1\dec{16.1} & 72.3\dec{18.3} \\  \hline

 \multirow{2}{*}{\textbf{SNIFFER}} 
   & Realistic & 84.3 & 78.4 & 93.2 & 85.1 & 92.3 & 76.2 & 83.5 \\
   & Diversified & 73.5 \dec{10.8} & 71.2 \dec{7.2} & 76.0 \dec{17.2} & 73.5 \dec{11.6} & 75.8 \dec{16.5} & 71.1 \dec{5.1} & 73.4 \dec{10.1} \\ \hline

 \multirow{2}{*}{\textbf{LEMMA}} 
   & Realistic & 79.6 & 73.4 & 92.8 & 82.0 & 90.2 & 66.5 & 76.5 \\
   & Diversified &  68.9 \dec{10.7} & 64.7 \dec{8.7} & 83.1 \dec{9.7} & 72.7 \dec{9.3} & 76.3 \dec{13.9} & 54.7 \dec{11.8} & 63.7 \dec{12.8} \\ \hline \hline

\end{tabular}
    \label{tab: main_results}
  \end{center}
\end{table*}

%% file: table/pollution_simple.tex
\begin{table}[t]
\renewcommand\arraystretch{1}
\setlength{\tabcolsep}{7 pt} 
\small
  \begin{center}
\begin{tabular}{ccl}
\hline 
\textbf{Infer Type} & \textbf{Data Type} & \textbf{Accuracy} \\ 
\hline

\multirow{2}{*}{\textbf{GPT-4o-mini}} & DI\_OT & 64.4  \\
& Polluted & 44.6 \dec{19.8}  \\ 
\hline

\multirow{2}{*}{\textbf{Claude-3.7-Sonnet}} & DI\_OT & 72.4  \\
& Polluted & 66.7 \dec{5.7}  \\ 
\hline

\multirow{2}{*}{\textbf{Qwen-VL}} & DI\_OT & 63.7   \\
& Polluted & 37.2 \dec{26.5}  \\ 
\hline

\multirow{2}{*}{\textbf{CMIE}} & DI\_OT & 72.4  \\
& Polluted & 66.8 \dec{5.6}  \\ 
\hline

\multirow{2}{*}{\textbf{SNIFFER}} & DI\_OT & 73.5   \\
& Polluted & 70.3 \dec{3.2} \\ 
\hline

\multirow{2}{*}{\textbf{LEMMA}} & DI\_OT & 68.9 \\
& Polluted & 41.9 \dec{27.0}  \\ 
\hline

\end{tabular}
\caption{\normalsize 
    Comparison of LVLM-based multimodal misinformation detection under malicious evidence contamination, evaluated on the DI\_OT setting with controlled news diversity.}
    \label{tab: pollution}
  \end{center}
\end{table}

%% file: appendix.tex
\setcounter{secnumdepth}{2} 

%



\appendix

\clearpage

\appendix

\section{Appendix}
\label{sec:appendix}

\subsection{Dataset Creation Details}
\label{app: Dataset_Creation_Details}

In section \ref{sec: data_creation}, we discussed the source data collection process of \textsc{DriftBench}. We outlines the key implementation details and the prompt templates used to generate the diversified and fabricated content in the \textsc{DriftBench}, including both text and image transformations.

\subsubsection{Text diversification (\textit{DT})}: Given a human written news text (\textit{OT}), we prompt GPT-4o \cite{DBLP:journals/corr/gpt-4o} to rephrase the original text while preserving its semantics.

\begin{tcolorbox}
[colback=black!2!white,colframe=white!50!black,boxrule=0.5mm]
You are given a news caption. Your task is to rewrite it while ensuring the original meaning remains unchanged. \\
News caption: \{\} \\
Your Response:
\end{tcolorbox}

\subsubsection{Text fabrications (\textit{FT})}: Given a human written news text (\textit{OT}), we prompt GPT-4o \cite{DBLP:journals/corr/gpt-4o} to generate a false narrative by altering the original news caption.

\begin{tcolorbox}
[colback=black!2!white,colframe=white!50!black,boxrule=0.5mm]
You are given a news caption. Your task is to rewrite the event described in the original caption with a false claim. \\
News caption: \{\} \\
Your Response:
\end{tcolorbox}

\subsubsection{Image diversification (\textit{DI})}: Given an authentic image (\textit{OI}), we utilize \textit{FLUX.1 Redux [dev]} \cite{labs2025flux1kontextflowmatching}, an open-source image-to-image variant generation model, to generate diversified image. We set the \textit{guidance\_scale} as 5.5 and \textit{num\_inference\_step} as 50.

\subsubsection{Image fabrications (\textit{FI})}: Given an false narrative (\textit{FI}), we utilize \textit{FLUX.1 [dev]} \cite{flux2024}, an open-source text-to-image generation model, to generate fabricated image. We set the \textit{guidance\_scale} as 3.5 and \textit{num\_inference\_step} as 50.

\begin{tcolorbox}
[colback=black!2!white,colframe=white!50!black,boxrule=0.5mm]
Generating an image based on the text description in a natural style. Text Description: [FT]
\end{tcolorbox}

\subsection{Human Evaluation}
\label{app: human_eval}
\begin{figure*}[t]
\begin{center}
    \includegraphics[width=1\linewidth]  {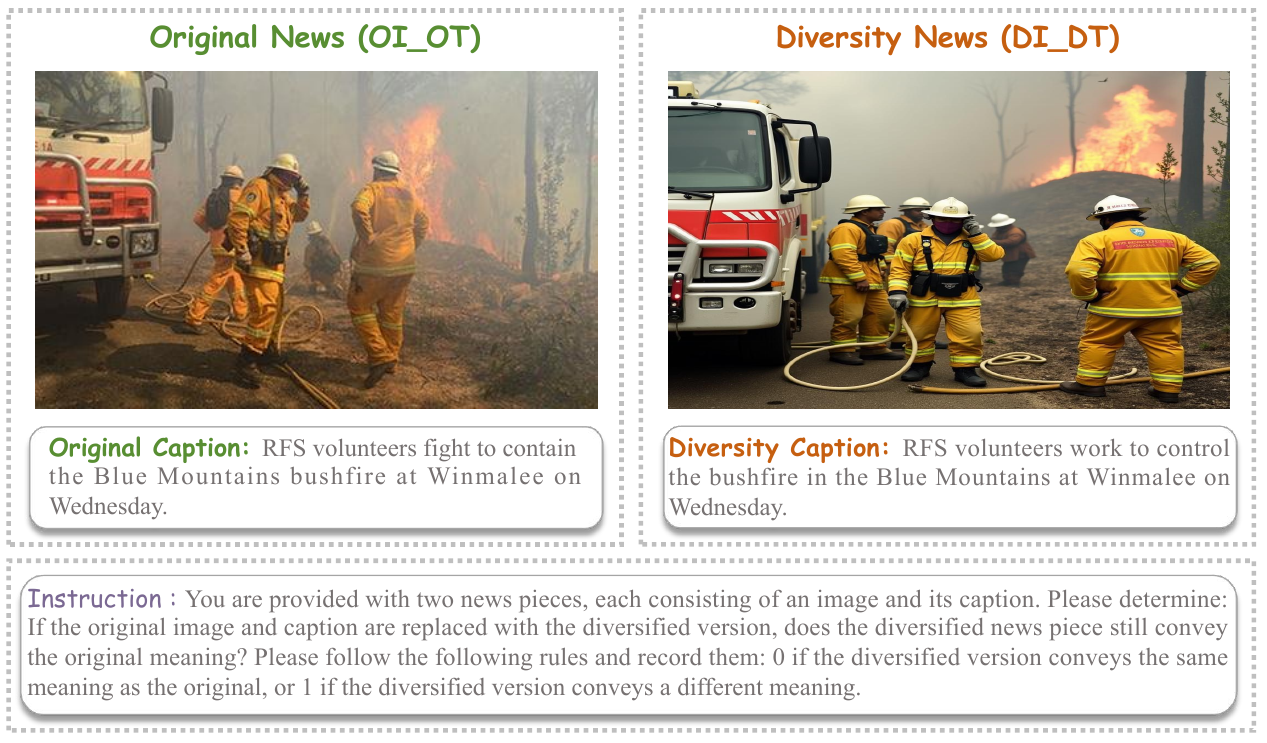}
    \caption{llustration of human evaluation instructions for validating the quality of DriftBench. We present the validation process of \textit{DI\_DT}, with the validation processes for \textit{DI\_OT} and \textit{OI\_DT} being similar.}
    \label{appfig:human_eval}
\end{center}
\end{figure*}

We illustrate the annotation process for human evaluation discussed in Section \ref{sec: human_eval} in Figure \ref{appfig:human_eval}.

\subsection{Malicious Evidence Contamination}
\label{app: malicious}
To implement malicious evidence contamination through guided retrieval, we prompt GPT-4o \cite{DBLP:journals/corr/gpt-4o} to generate evidence with malicious intent, such as evidence supporting or refuting news instances, and directly insert it into the original evidence list. The original evidence list is curated by crawling data from the web and selecting the top 5 pieces of evidence. To evaluate the impact of evidence contamination, we insert the generated malicious evidence into the list and use it for model inference.

\begin{tcolorbox}
[colback=black!2!white,colframe=white!50!black,boxrule=0.5mm]
You are given an news image and corresponding news caption. 
 Your task is to rewrite this description into a \textbf{supporting/refuting} event — that is, to \textbf{support/refute} the original caption. \\
Original caption: \{\} \\
 Your response:
\end{tcolorbox}

\subsection{Error Attribution}
\label{app: error_attribution}
We listed detailed error attribution taxonomy in Figure \ref{appfigure:error_definition} discussed in Section \ref{sec: eval_task_setup}. 

Error attribution identifies two types of errors caused by two levels of drift: model misperception drift errors and evidence drift errors. Model misperception drift errors are directly caused by diversified input, leading to model misclassification, and are unrelated to external evidence. Evidence drift errors occur when the model’s misclassification is triggered by changes in the external retrieved evidence following diversification.

\begin{figure*}[!ht]
    \begin{tcolorbox}[colback=black!2!white, colframe=white!50!black, boxrule=0.5mm]
    \textbf{Model misperception drift errors: }
    \begin{itemize}
        \item \textbf{Text–Image inconsistency}: The image and text do not align, leading to contradictions, such as the wrong person, place, action, or object being depicted.
        \item \textbf{Text–Image ambiguous or weak match}: The image is too vague or the caption is too generic, making it difficult to establish a clear and meaningful connection between them.
        \item \textbf{Image and caption match, but the claim itself is fabricated}: The image and caption are consistent, but the event or claim described is fabricated or never occurred (e.g., a historical event that never happened).
    \end{itemize}

   \textbf{Model misperception drift errors: }
    \begin{itemize}
           \item \textbf{No supporting evidence found}: The retrieval process fails to find any relevant evidence or retrieves evidence from irrelevant domains, leading to a lack of support for the claim.
            \item \textbf{Misleading retrieved evidence}: The retrieved evidence may have similar names or topics to the claim but is ultimately a poor or incorrect match.
            \item \textbf{Mismatch in time or location between the evidence and the caption}: The time or location of the retrieved evidence does not align with that presented in the caption, leading to inconsistencies.
            \item \textbf{Over-reliance on low-quality or noisy evidence}: The model excessively depends on low-quality or noisy evidence, which undermines the overall reliability and validity of the claim's verification.
    \end{itemize}
    
    \end{tcolorbox}
    \caption{Error attribution taxonomy according to model-level misperception drift and evidence-level drift.}
    \label{appfigure:error_definition}
\end{figure*}

\subsection{Explanation Evaluation}
\label{app: explanation_eval}
\begin{figure*}[h]
\begin{center}
    \includegraphics[width=1\linewidth]  {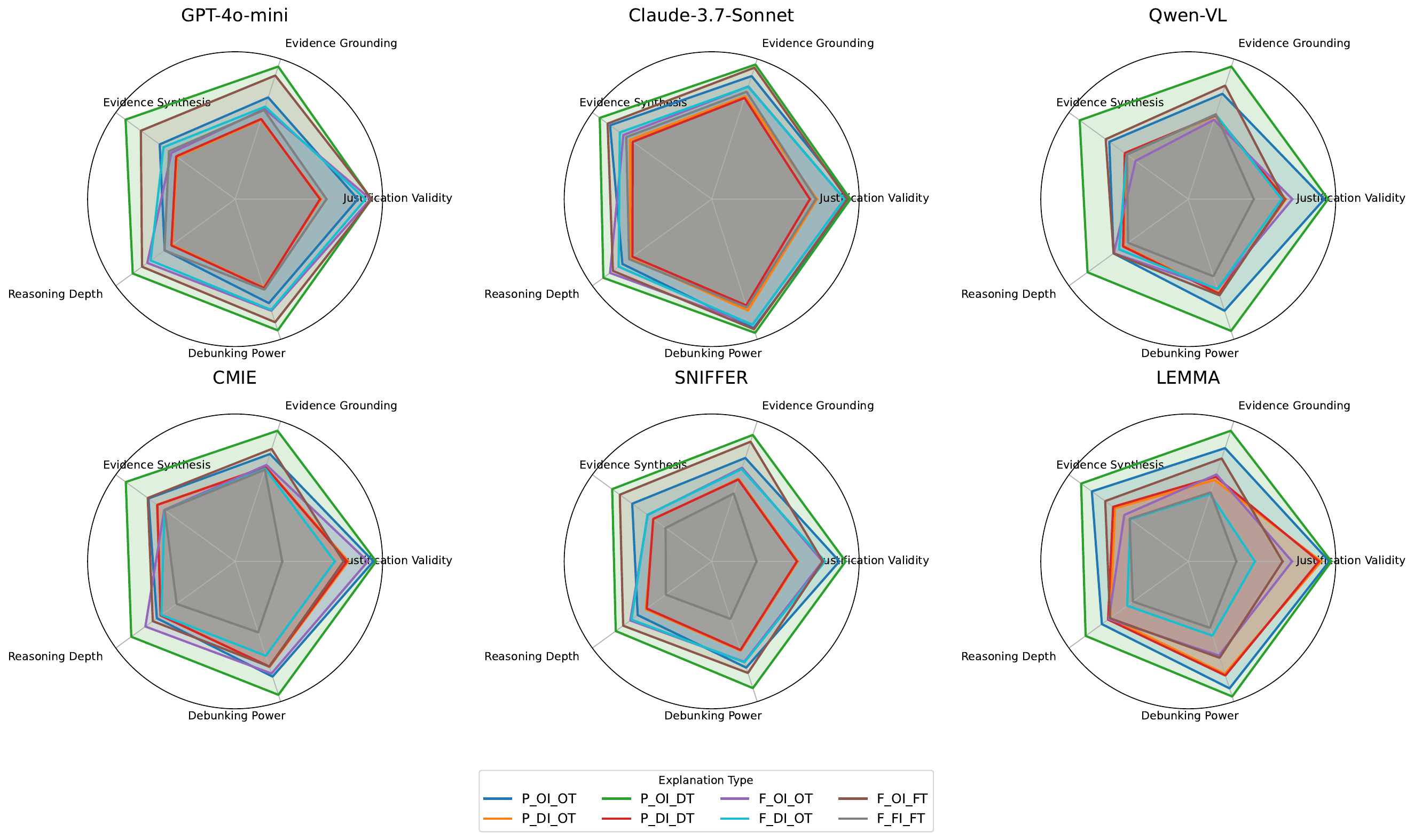}
    \caption{Model reasoning capabilities under different dimension.}
    \label{appfig:Radar_all}
\end{center}
\end{figure*}

To analyze the reasoning behavior of the model, we evaluated the explanations generated by different methods discussed in \ref{sec: eval_task_setup}. The detailed evaluation dimensions and criteria is shown in Figure \ref{appfigure: explanation_definition}, and the results for all models are shown in the Figure \ref{appfig:Radar_all}. In this part, we also use GPT-4o \cite{DBLP:journals/corr/gpt-4o} to conduct our experiment.

\begin{figure*}[!ht]
    \begin{tcolorbox}[colback=black!2!white, colframe=white!50!black, boxrule=0.5mm]
    \centering
    \begin{itemize}
        \item \textbf{Justification Validity (1–5):} Based on the Ground Truth Label, how well does explanation support the correct label? \\
    \textbf{- Score 5:} Justification is completely sound and logical. \\
    \textbf{- Score 1:} Justification is flawed or supports the wrong conclusion.

    \item \textbf{Evidence Grounding (1–5):} Does explanation explicitly and accurately attribute its claims to the specific (drifted) evidence? \\
    \textbf{- Score 5:} Each key argument is precisely linked to a specific piece of evidence. \\
    \textbf{- Score 1:} Makes claims without any reference to the evidence.

    \item  \textbf{Evidence Synthesis (1–5):} Does explanation weave multiple points from evidence into a coherent argument? \\
    \textbf{- Score 5:} Skillfully combines multiple evidence points to form a robust conclusion. \\
    \textbf{- Score 1:} Fails to handle multiple evidence points, leading to confusion.

    \item  \textbf{Reasoning Depth (1–5):} Does the reasoning go beyond surface-level matching? \\
    \textbf{- Score 5:} Uncovers deep, underlying consistencies or inconsistencies. \\
    \textbf{- Score 1:} Reasoning is superficial and based on simple keyword matching.

    \item   \textbf{Debunking Power (1–5):} How effective would explanation be at convincing a layperson of the ground truth? \\
    \textbf{- Score 5:} Highly persuasive, clear, and provides helpful context. \\
    \textbf{- Score 1:} Vague, unconfident, or too brief to be persuasive.
    
    \end{itemize}
    \end{tcolorbox}
    \caption{Five evaluation dimensions for explanations generated by MMD method.}
    \label{appfigure: explanation_definition}
\end{figure*}

\subsection{Detailed Results}
\label{app: detailed_results}
We present the detailed performance of \textit{Real vs. Fake} instances across all diversified types discussed in Section \ref{sec: Performance_under_GenAI-Driven_Diversity} in Table \ref{apptable: all_types}. Additionally, we report the results of robustness to evidence contamination in Table \ref{table: pollution_all}, as discussed in Section \ref{sec: robustness_to_evidence_contamination}.

\subsection{Evaluation Prompts}
\label{app: evaluation_prompts}
We present our prompts for evaluating Vanilla LVLMs in Figure \ref{appfig:LVLMs-prompt}
.

\input{table/merge_tab} 
\input{table/pollution}

\input{prompts/vanilla_LVLMs}

%% file: table/merge_tab.tex
\begin{table*}[t]
\renewcommand\arraystretch{1.0}
\setlength{\tabcolsep}{5 pt}
\small
  \begin{center}
    \caption {Detailed comparison of LVLM-based MMD methods under the all types diversification.}
    \begin{tabular}{c c l l l l}
    \hline \hline
    \multirow{2}{*}{\textbf{Infer Type}} & \multirow{2}{*}{\textbf{Data Type}} & \multicolumn{2}{c}{\textbf{Real}} & \multicolumn{2}{c}{\textbf{Fake}} \\ \cline{3-6}
    & & \multicolumn{1}{c}{\textbf{Recall}} & \multicolumn{1}{c}{\textbf{F1}} & \multicolumn{1}{c}{\textbf{Recall}} & \multicolumn{1}{c}{\textbf{F1}} \\ \hline

    \multirow{4}{*}{\textbf{GPT-4o-mini}} 
      & OI\_OT & 75.6 & 86.1 & 91.1 & 95.3 \\
      & DI\_OT & 35.6 \dec{40.0} & 52.6 \dec{33.5} & 93.1 \imp{2.0} & 96.4 \imp{1.1} \\
      & OI\_DT & 74.7 \dec{0.9} & 85.5 \dec{0.6} & 94.5 \imp{3.4} & 97.1 \imp{1.8} \\
      & DI\_DT & 34.7 \dec{40.9} & 51.5 \dec{34.6} & 64.2 \dec{26.9} & 78.2 \dec{17.1} \\
    \hline

    \multirow{4}{*}{\textbf{Claude-3.7-Sonnet}} 
      & OI\_OT & 89.2 & 94.3 & 87.5 & 93.3 \\
      & DI\_OT & 53.5 \dec{35.7} & 69.7 \dec{24.6} & 91.4 \dec{35.7} & 95.5 \dec{24.6} \\
      & OI\_DT & 83.7 \dec{5.5} & 91.2 \dec{3.1} & 94.5 \dec{5.5} & 97.1 \dec{3.1} \\
      & DI\_DT & 45.4 \dec{43.8} & 62.5 \dec{31.8} & 75.5 \dec{43.8} & 86.1 \dec{31.8} \\
    \hline

    \multirow{4}{*}{\textbf{Qwen-VL}} 
      & OI\_OT & 91.1 & 95.3 & 56.3 & 72.1 \\
      & DI\_OT & 62.1 \dec{28.4} & 77.1 \dec{18.2} & 64.7 \dec{78.6} & 78.6 \dec{24.6} \\
      & OI\_DT & 89.2 \dec{1.9} & 94.3 \dec{1.0} & 59.8 \dec{5.5} & 74.8 \dec{3.1} \\
      & DI\_DT & 61.9 \dec{29.2} & 74.5 \dec{20.8} & 45.7 \dec{43.8} & 62.8 \dec{31.8} \\
    \hline

    \multirow{4}{*}{\textbf{CMIE}} 
      & OI\_OT & 93.6 & 96.7 & 88.2 & 93.7 \\
      & DI\_OT & 72.8 \dec{20.8} & 84.2 \dec{12.5} & 72.1 \dec{16.1} & 83.8 \dec{9.9} \\
      & OI\_DT & 93.7 \imp{0.1} & 96.8 \imp{0.1} & 75.1 \dec{13.1} & 85.5 \dec{8.2} \\
      & DI\_DT & 72.4 \dec{21.2} & 83.9 \dec{12.8} & 23.1 \dec{65.1} & 37.5 \dec{57.8} \\
    \hline

    \multirow{4}{*}{\textbf{SNIFFER}} 
      & OI\_OT & 93.2 & 96.5 & 76.2 & 86.5 \\
      & DI\_OT & 76.0 \dec{17.2} & 86.4 \dec{10.1} & 71.1 \dec{5.1} & 83.1 \dec{3.4} \\
      & OI\_DT & 93.4 \imp{0.2} & 96.6 \imp{0.1} & 78.3 \imp{2.1} & 87.8 \imp{1.3} \\
      & DI\_DT & 74.8 \dec{18.4} & 85.6 \dec{10.9} & 17.9 \dec{0.0} & 30.3 \dec{0.0} \\
    \hline

    \multirow{4}{*}{\textbf{LEMMA}} 
      & OI\_OT & 92.8 & 96.2 & 77.5 & 79.8 \\
      & DI\_OT & 83.1 \dec{9.7} & 90.7 \dec{5.5} & 54.7 \dec{22.8} & 70.7 \dec{9.1} \\
      & OI\_DT & 92.0 \dec{0.8} & 95.8 \dec{0.4} & 73.8 \dec{3.7} & 84.9 \imp{5.1} \\
      & DI\_DT & 82.8 \dec{10.0} & 90.6 \dec{5.6} & 25.5 \dec{65.6} & 40.6 \dec{54.7} \\
    \hline \hline
    \end{tabular}
    \label{apptable: all_types}
  \end{center}
\end{table*}

%% file: table/pollution.tex
\begin{table*}[t]
\renewcommand\arraystretch{1}
\setlength{\tabcolsep}{3 pt} 
\small
  \begin{center}
    \caption{Detailed comparison of LVLMs-based MMD performance with malicious evidence contamination under the \textit{DI\_OT} controlled news diversity.}
\begin{tabular}{cclllllll}

\hline \hline
\multirow{2}{*}{\textbf{Infer Type}} & \multirow{2}{*}{\textbf{Data Type}} & \multirow{2}{*}{\textbf{Accuracy}} &  \multicolumn{3}{c}{\textbf{Real}} & \multicolumn{3}{c}{\textbf{Fake}} \\ \cline{4-9}
  & & & \multicolumn{1}{c}{\textbf{Precision}} & \multicolumn{1}{c}{\textbf{Recall}} & \multicolumn{1}{c}{\textbf{F1}} & \multicolumn{1}{c}{\textbf{Precision}} & \multicolumn{1}{c}{\textbf{Recall}} & \multicolumn{1}{c}{\textbf{F1}} \\ \hline

  \multirow{2}{*}{\textbf{GPT-4o-mini}} & DI\_OT & 64.4 & 83.7 & 35.7 & 50.0 & 59.1 & 93.1 & 72.3 \\
  & Polluted & 44.6 \dec{19.8} & 18.8 \dec{64.9} & 3.3 \dec{32.4} & 5.5 \dec{44.5} & 47.1 \dec{12.0} & 86.0 \dec{7.1} & 60.8 \dec{11.5}\\ \hline

  \multirow{2}{*}{\textbf{Claude-3.7-Sonnet}} & DI\_OT & 72.4 & 86.1 & 53.5 & 66.0 & 66.3 & 91.4 & 76.8 \\
  & Polluted & 66.7 \dec{5.7} & 79.0 \dec{7.1} & 45.4 \dec{8.1} & 57.7 \dec{8.3} & 61.7 \dec{4.6} & 87.9 \dec{3.5} & 72.5 \dec{4.3}\\ \hline

    \multirow{2}{*}{\textbf{Qwen-VL}} & DI\_OT & 63.7 & 63.9 & 62.7 & 63.3 & 63.5 & 64.7 & 64.1 \\
  & Polluted & 37.2 \dec{26.5} & 32.3 \dec{31.6} & 23.2 \dec{39.5} & 27.0 \dec{36.3} & 39.9 \dec{23.6} & 51.2 \dec{13.5} & 44.8 \dec{19.3}\\ \hline

   \multirow{2}{*}{\textbf{CMIE}} & DI\_OT & 72.4 & 72.2 & 72.8 & 72.5& 72.6 & 72.1 & 72.3 \\
  & Polluted & 66.8 \dec{5.6} & 66.9 \dec{5.3} & 66.2 \dec{6.6} & 66.6 \dec{5.9} & 66.6 \dec{6.0} & 67.3\dec{4.8} & 66.9 \dec{5.4}\\ \hline

     \multirow{2}{*}{\textbf{ SNIFFER}} & DI\_OT & 73.5 & 71.2 & 76.0 & 73.5& 75.8 & 71.1 & 73.4 \\
  & Polluted & 70.3 \dec{3.2} & 70.2 \dec{1.0} & 73.4 \dec{2.6} & 71.8 \dec{1.7} & 70.4 \dec{5.4}& 67.1\dec{4.0} & 68.7 \dec{4.7}
  
  \\  \hline
  
     \multirow{2}{*}{\textbf{LEMMA}} & DI\_OT & 68.9 & 64.7 & 83.1 & 72.7& 76.3 & 54.7 & 63.7 \\
  & Polluted & 41.9 \dec{27.0}& 42.1 \dec{22.6} & 43.2 \dec{39.9} & 42.7 \dec{30.0} & 41.7 \dec{34.6} & 40.7 \dec{14.0} & 41.2 \dec{22.5}
  
  \\  \hline
 \hline

\end{tabular}
    \label{table: pollution_all}
  \end{center}
\end{table*}

%% file: prompts/vanilla_LVLMs.tex
\begin{figure*}[t]
\begin{prompt}{Misinformation Detection Using Vanilla LVLMs}
\small  
You are given a news post, inncluding an image, a piece of text, and a list of titles in which the image has appeared. Your task is to predict whether there is misinformation between the given image and text.

Generate a JSON object with two properties: 'label', 'explanation'. 
The return value of 'label' property should be selected from ["real", "fake"].
Fake indicates there is misinformation between the given image and text.
Real indicates that there is no misinformation between the given image and text.
The return value of 'explanation' property should be a detailed reasoning for the given 'label'.

The given text: \{\}

The list of titles related to the image content: \{\}

Your Response:

\end{prompt}
\caption{Prompt of misinformation detection using Vanilla LVLMs.}
\label{appfig:LVLMs-prompt}
\end{figure*}